\pdfoutput=1

\documentclass[11pt]{article}

\usepackage{ACL2023}

\usepackage{times}
\usepackage{latexsym}

\usepackage[T1]{fontenc}

\usepackage[utf8]{inputenc}

\usepackage{microtype}

\usepackage{caption}

\usepackage{inconsolata}
\usepackage{times}
\usepackage{latexsym}

\usepackage{longtable}
\usepackage{tabu}
\usepackage{arydshln}

\usepackage{xspace}
\usepackage{graphicx}
\usepackage{wrapfig}
\usepackage{CJK,algorithm,algorithmic,amssymb,amsmath,array,epsfig,graphics,float,subfigure,verbatim,epstopdf}
\usepackage{enumitem}
\usepackage{color,soul}
\usepackage{hhline}
\usepackage{multirow}
\usepackage{multicol}
\usepackage{xcolor}
\usepackage{hyperref}
\usepackage{cleveref}

\usepackage{xurl}

\usepackage{tabularx}
\usepackage{bm}
\usepackage{seqsplit}
\usepackage{stmaryrd}
\usepackage{booktabs}

\usepackage{multirow}
\newcommand{\thickhline}{\noalign{\hrule height 2pt}}

\NewDocumentCommand{\heng}
{ mO{} }{\textcolor{red}{\textsuperscript{\textit{Heng}}\textsf{\textbf{\small[#1]}}}}

\NewDocumentCommand{\carl}
{ mO{} }{\textcolor{blue}{\textsuperscript{\textit{Carl}}\textsf{\textbf{\small[#1]}}}}

\title{\textit{L+M-24}: Building a Dataset for Language+Molecules @ ACL 2024}

\author{Carl Edwards$^1$, Qingyun Wang$^1$, Lawrence Zhao$^2$ \and Heng Ji$^1$ \\
  $^1$University of Illinois Urbana-Champaign $^2$Yale University \\
  \texttt{\{cne2, qingyun4, hengji\}@illinois.edu, larry.zhao@yale.edu }}

\usepackage{amsmath}
\usepackage{tabularx}
\usepackage{graphicx}

\newcommand{\name}{$\text{\textit{L+M-24}}$}

\begin{document}
\maketitle
\begin{abstract}
Language-molecule models have emerged as an exciting direction for molecular discovery and understanding. However, training these models is challenging due to the scarcity of molecule-language pair datasets. At this point, datasets have been released which are 1) small and scraped from existing databases, 2) large but noisy and constructed by performing entity linking on the scientific literature, and 3) built by converting property prediction datasets to natural language using templates. In this document, we detail the \name{} dataset, which has been created for the Language + Molecules Workshop shared task at ACL 2024. In particular, \name{} is designed to focus on three key benefits of natural language in molecule design: compositionality, functionality, and abstraction.\footnote{The dataset, finetuned baseline, and evaluation code are released publicly at \href{https://github.com/language-plus-molecules/LPM-24-Dataset}{github.com/language-plus-molecules/LPM-24-Dataset} through \href{https://huggingface.co/language-plus-molecules}{HuggingFace}.}

\end{abstract}

\section{Introduction}

The world faces an enormous number of problems in the coming decades on scales of complexity never-before-seen, in areas such as climate change, healthcare, and pandemics. To address these issues, we need to discover inventive scientific solutions which are scalable, flexible, and inexpensive. Broadly speaking, many of these problems will require molecular solutions from the chemistry domain, such as developing new drugs (e.g. kinase inhibitors \cite{ferguson2018kinase}), materials (e.g. organic photovoltaics \cite{kippelen2009organic}), and chemical processes \cite{reactionminer2023}. %
These solutions exist in extremely large search spaces, which makes AI tools a necessity. 

Language-molecule models have emerged as an exciting direction for molecular discovery and understanding \cite{edwards2021text2mol, zeng-etal-2022-deep, edwards2022translation, su2022molecular, liu2022multi, xu2023multilingual, christofidellis2023unifying, liu2023molxpt, luo2023molfm, zhao2023adversarial, seidl2023enhancing}. However, training these models is challenging due to the scarcity of molecule-language pair datasets. At this point, datasets have been released which are 1) small and scraped from existing databases \cite{edwards2021text2mol, zeng2023interactive, liu2023git, liu2023molca, pei2023biot5}, 2) large but noisy and constructed by performing entity linking on the scientific literature \cite{zeng-etal-2022-deep, su2022molecular}, and 3) template-based built on prediction datasets \cite{zhao2023gimlet, fang2023mol}. Approaches utilizing pseudo-data have also been attempted \cite{chen2023artificially}. These approaches have helped remedy the problem of data scarcity in this domain; however, these approaches frequently ignore key benefits of natural language: 1) compositionality, 2) abstraction, and 3) functionality \cite{zhang2023artificial}. To this end, for the Language + Molecules Workshop at ACL 2024, we release \name{}, which we construct to test these three goals, particularly compositionality, using recently released data sources \cite{zhao2023scientific, kosonocky2023mining, wishart2023chemfont}. \name{} is divided into four categories with important applications in the small-molecule domain: 1) Biomedical, 2) Light and Electricity, 3) Human Interaction and Organoleptics, and 4) Agriculture and Industry. Improving understanding of these applications can have important implications in problems such as drug discovery, climate issues, more efficient and green industrial processes, and improved food production.%

\section{Task Formulation}
The dataset is primarily intended for language$\leftrightarrow$molecule translation, which consists of two tasks: generating 1) a caption given a molecule and 2) a molecule given a description.

\subsection{Designing for Compositionality, Abstraction, and Function}
Overall, we focused on four primary categories of importance: 1) Biomedical, 2) Light and Electricity, 3) Human Interaction and Organoleptics, and 4) Agriculture and Industry. These categories and three properties from each are displayed in Table \ref{tab:accents}. The biomedical category is focused on drug properties, functions, and interaction with proteins. Light and electricity is focused on the ability for a molecule to produce or absorb light or electricity. Human interaction and organoleptics focuses on the effect and experience molecules cause in humans. Agriculture and industry focuses on molecules used in industrial processes and food production. 

Based on our data sources (below), the properties we have selected already encode a large degree of functionality, enhanced by our manual curation. Further, since these properties are generally short phrases indicating functionality, they are also abstract and apply to many molecules  (e.g., ``insecticide''). For compositionality, we explicitly select certain pairs of properties which we hold out of the dataset. For example, a molecule may share two properties which are desirable together (e.g., low toxicity and fungicidal). \name{} will help to evaluate whether model's can generalize to unseen compositions of properties.

\section{Data Sources}

We constructed our dataset using three different databases. We will first describe the process we used to extract information from each, followed by our overall strategy for adding hierarchy into the dataset. We want to deeply thank the authors of these resources for making them publicly available for the community.

\subsection{PubChem}

We used properties extracted from PubChem \cite{kim2016pubchem, kim2019pubchem} as described in \cite{zhao2023adversarial}. Properties from this approach include odor, taste, and decomposition. We note these properties consist of molecule-specific descriptions, which the other data sources do not provide. 

\subsection{Chemical Function (CheF)}

Here, we used functional properties extracted from patent literature by \citet{kosonocky2023mining}. This allowed us to capture molecules from the patent literature in addition to the scientific literature. Here, we started with CheF prefinal\_v3\footnote{obtained via personal communication.}. We created a set of properties from both CheF's property summarizations and from the ChatGPT summarization source. For the summarization source, we also applied the WordNet lemmatizer \cite{bird2009natural} for deduplication. After obtaining a list of properties, we removed properties pertaining to less than 100 molecules. We then kept properties falling into the categories of ``X-icide'', ``anti-X'', ``X treatment'', ``X modulators'', ``X inhibitors'', ``X agonists'', ``X antagonists'', ``light'', and ``electricity.'' We manually removed uninformative labels which were too broad or didn't describe enough function. Further, we manually corrected errors in label naming and duplication. 

\subsection{ChemFOnt: the chemical functional ontology resource}

In addition to CheF, we also take advantage of another new chemical function data resource: ChemFOnt \cite{wishart2023chemfont}. From this datasource, we collect three categories: health effect relations, organoleptic effect relations, and role relations. %

\begin{table}
\centering
\scalebox{0.82}{
\begin{tabular}{c}
\hline
\textbf{Biomedical} \\
\hline
anti neoplastic \\
glaucoma treatment \\
capillarigenic \\
\hline 
\textbf{Light and Electricity} \\
\hline
photoelectric conversion \\
photopolymerization \\
dielectric \\
\hline
\end{tabular} 
}
\scalebox{0.82}{
\begin{tabular}{c}
\hline
\textbf{Human Interaction} \\
\hline
pungent \\
bitter \\
nephrotoxic \\
\hline
\textbf{Agriculture and Industry} \\
\hline
herbicide \\
emulsifier \\
carcinogen \\
\hline
\end{tabular}
}
\caption{Example properties in the dataset. Antineoplastic drugs are used to treat cancer. Glaucoma is a group of eye diseases. Capillarigenic means producing or causing capillaries. Pungent means having a strong taste or smell. Nephrotoxic is toxicity in the kidneys. Photoelectric conversion is the conversion of light into electricity. Photopolymerization is the process through which monomers are linked together through a photochemical reaction. A dielectric is a poor conductor of electricity but can be polarized. A herbidicde is toxic to plants. An emulsifier stabilizes an emulsion. A carcinogen is an agent capable of causing cancer. The full property list and number of occurrences is available in the online data repository.}
\label{tab:accents}
\end{table}

\begin{figure*}[h!]
  \centering
  \includegraphics[width=1.0\textwidth]{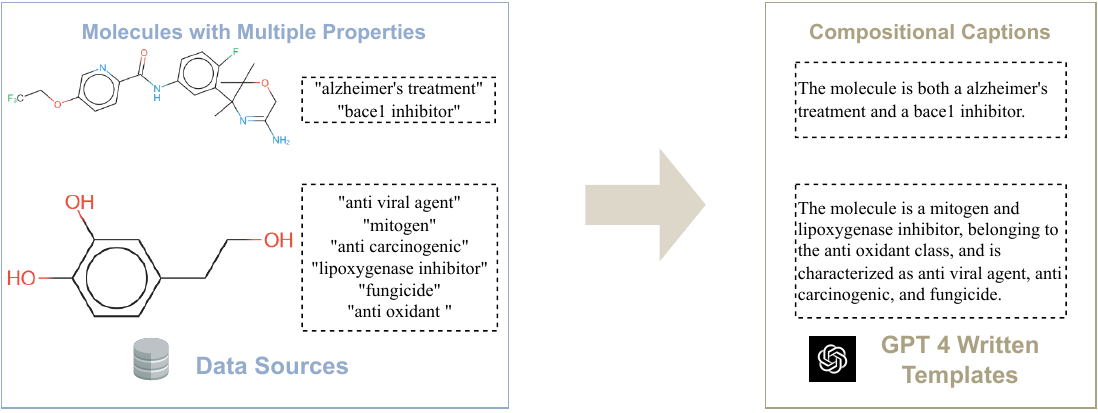}
\caption{Example descriptions created for molecules from the training set.}
\label{fig:challenge_and_approach}
\end{figure*}
\section{Dataset Details}
To convert these properties to natural language, we follow a template-based procedure using GPT-4 \cite{openai2023gpt4} generated compositional templates. 

\subsection{Template Generation}

We utilize GPT-4~\cite{openai2023gpt4} to generate specific templates for each combinations of molecular properties. Specifically, we manually write six templates:  ``\textit{The molecule is a <0>.}''; 
``\textit{It belongs to the <1> class of molecules.}'';
``\textit{It has an effect on <2>.}'';
``\textit{It impacts <3>.}'';
``\textit{The molecule is <4>.}'';
and ``\textit{The molecule has a <5>.}''. Subsequently, we use GPT-4 to generate a unique sentence template for each possible combination by rephrasing up to six combinations of the six initial templates as a single sentence. Ultimately, this process results in the generation of 917 distinct templates. The templates were manually checked and corrected to have a matching standard. The prompts and in-context examples for GPT-4 are given in the Appendix.

\subsection{Converting Templates to Descriptions}
For all properties in \name{}, we first assigned them to possible templates based on their category or by individual consideration. Certain properties (e.g., polymerization, decomposition) were expressed in sentence format, so we did not use templates. Given a molecule with $n$ properties, we first looked for a template that had the correct slots (e.g., <0>, <2>, and <2>) for its properties. When we found possible templates, we picked one at random and used it to generate a sentence for the molecule's properties. 
If there were no matching templates, we split the properties into two separate equal-sized groups and tried with each group. We return the concatenation of the two sentence templates as the molecule description. Note this process can repeat multiple times. 

We note that we are also releasing a version of the dataset with 5 captions for each molecule. In this case, we split group sizes at random. Further, we split sentences apart 50\% of the time (even when there were matching templates) to increase caption diversity. 

\begin{table*}[ht!]
\resizebox{\textwidth}{!}{
\centering
\tiny
\begin{tabular}{ c|ccccccc }

\multicolumn{1}{c}{\textbf{Model}} & \multicolumn{1}{c}{BLEU-2} & \multicolumn{1}{c}{BLEU-4} & \multicolumn{1}{c}{ROUGE-1} & \multicolumn{1}{c}{ROUGE-2} & \multicolumn{1}{c}{ROUGE-L} & \multicolumn{1}{c}{METEOR} & \multicolumn{1}{c}{Text2Mol} \\
\thickhline
Ground Truth& &  &  &  &  &  & 11.30 \\\hline
MolT5-Small & 70.9 & 51.2 & 74.5 & 55.8 & 54.4 & 70.1 & 10.79 \\
MolT5-Base & 73.8 & 53.5 & 75.0 & 55.9 & 53.9 & 71.8 & 8.53 \\
MolT5-Large & 76.9 & 55.6 & 77.7 & 58.0 & 55.7 & 74.3 & 10.06 \\
Meditron-7B & 79.2 & 57.6 & 79.7 & 60.2 & 57.5 & 75.7 & 11.91  \\
\end{tabular}
}
\caption{Molecule captioning results on the validation split of \name{}. Rouge scores are F1 values.}
\label{tab:results_captioning}
\end{table*}

\begin{table*}[ht!]
\resizebox{\textwidth}{!}{
\centering
\tiny
\begin{tabular}{ c|ccccccccccccccc|ccc }

 & \multicolumn{3}{c}{Overall} & \multicolumn{3}{c}{Biomedical} & \multicolumn{3}{c}{Light+Electro} & \multicolumn{3}{c}{Human Interaction} & \multicolumn{3}{c}{Agr.+Industry} & \multicolumn{3}{c}{Held-out Combos} \\
\multicolumn{1}{c|}{\textbf{Model}} & P & R & F-1 & P & R & F-1 & P & R  & F-1 & P & R  & F-1 & P & R & F-1 & P & R & F-1  \\
\thickhline
MolT5-Small & 84.83 & 8.24 & 7.88 & 85.13 & 23.23 & 23.33 & 62.42 & 4.85 & 3.27 & 96.77 & 0.57 & 0.56 & 95.00 & 4.32 & 4.36 & 0.00 & 0.00 & 0.00 \\
MolT5-Base & 64.11 & 9.94 & 9.46 & 79.58 & 23.89 & 24.02 & 16.08 & 5.82 & 3.36 & 63.94 & 5.01 & 5.18 & 96.85 & 5.05 & 5.27 & 0.00 & 0.00 & 0.00 \\
MolT5-Large & 59.57 & 12.49 & 11.71 & 70.27 & 26.99 & 26.87 & 16.96 & 10.90 & 7.39 & 62.77 & 5.99 & 6.27 & 88.29 & 6.06 & 6.31 & 0.00 & 0.00 & 0.00 \\
Meditron-7B & 33.60 & 16.33 & 16.81 & 57.19 & 33.96 & 35.27 & 26.51 & 16.48 & 17.49 & 29.54 & 7.52 & 7.07 & 21.18 & 7.35 & 7.40 & 12.35 & 0.29 & 0.56 \\
\end{tabular}
}
\caption{Property-specific molecule captioning results on the validation split of \name{}.}
\label{tab:prop_results_captioning}
\end{table*}

\begin{table*}[ht!]
\resizebox{\textwidth}{!}{
\centering
\tiny
\begin{tabular}{ c|cccccccccccccccccc }

\multicolumn{1}{c|}{\textbf{Model}} & P & R & F-1 & P & R & F-1 & P & R  & F-1 & P & R  & F-1 & P & R & F-1 & P & R & F-1  \\
\thickhline
 & \multicolumn{3}{c}{\textbf{X-icides}} & \multicolumn{3}{c}{\textbf{Toxins}} & \multicolumn{3}{c}{\textbf{Light}} & \multicolumn{3}{c}{\textbf{Electricity}} & \multicolumn{3}{c}{\textbf{X-inhibitors}} & \multicolumn{3}{c}{\textbf{anti-X}} \\
MolT5-Small & 100.00 & 0.00 & 0.00 & 100.00 & 0.00 & 0.00 & 24.85 & 9.69 & 6.54 & 100.00 & 0.00 & 0.00 & 3.42 & 0.43 & 0.09 & 1.96 & 0.00 & 0.00 \\
MolT5-Base & 100.00 & 0.00 & 0.00 & 67.45 & 8.51 & 8.84 & 28.00 & 11.51 & 6.52 & 4.17 & 0.12 & 0.20 & 2.20 & 0.58 & 0.11 & 9.70 & 0.23 & 0.15 \\
MolT5-Large & 100.00 & 0.00 & 0.00 & 69.42 & 10.29 & 10.85 & 15.77 & 12.28 & 8.16 & 18.14 & 9.52 & 6.62 & 8.90 & 2.28 & 1.13 & 4.32 & 1.16 & 0.61 \\
Meditron-7B & 100.00 & 0.00 & 0.00 & 48.79 & 11.75 & 11.05 & 29.10 & 20.64 & 20.64 & 23.93 & 12.33 & 14.34 & 35.69 & 19.91 & 22.65 & 14.79 & 9.34 & 8.98 \\
\hline
 & \multicolumn{3}{c}{\textbf{X-modulator}} & \multicolumn{3}{c}{\textbf{X-agonist}} & \multicolumn{3}{c}{\textbf{X-antagonist}} & \multicolumn{3}{c}{\textbf{X-treatment}} & \multicolumn{3}{c}{\textbf{X-disease}} & \multicolumn{3}{c}{\textbf{X cancer}} \\
MolT5-Small & 100.00 & 0.00 & 0.00 & 100.00 & 0.00 & 0.00 & 100.00 & 0.00 & 0.00 & 55.49 & 1.99 & 1.70 & 87.44 & 50.08 & 49.94 & 71.86 & 21.03 & 24.27 \\
MolT5-Base & 100.00 & 0.00 & 0.00 & 100.00 & 0.00 & 0.00 & 100.00 & 0.00 & 0.00 & 58.90 & 2.25 & 1.80 & 94.61 & 55.16 & 59.18 & 45.06 & 25.49 & 24.54 \\
MolT5-Large & 21.30 & 0.58 & 0.88 & 5.91 & 1.96 & 1.23 & 14.30 & 0.58 & 0.42 & 14.27 & 2.67 & 2.22 & 97.18 & 81.07 & 81.86 & 65.76 & 52.06 & 51.56 \\
Meditron-7B & 42.43 & 21.24 & 24.98 & 39.19 & 23.23 & 26.35 & 34.22 & 18.98 & 21.15 & 28.75 & 11.35 & 15.13 & 97.34 & 81.11 & 82.02 & 79.80 & 68.65 & 72.62 \\
\end{tabular}
}
\caption{Selected subproperty group-specific molecule captioning results on the validation split of \name{}.}
\label{tab:subprop_results_captioning}
\end{table*}

\begin{table*}[ht!]
\resizebox{\textwidth}{!}{
\centering
\begin{tabular}{ c|ccccccccc }

\multicolumn{1}{c}{\textbf{Model}} & \multicolumn{1}{c}{BLEU$\uparrow$} & \multicolumn{1}{c}{Exact$\uparrow$} & \multicolumn{1}{c}{Levenshtein$\downarrow$} & \multicolumn{1}{c}{MACCS FTS$\uparrow$} & \multicolumn{1}{c}{RDK FTS$\uparrow$} & \multicolumn{1}{c}{Morgan FTS$\uparrow$} & \multicolumn{1}{c}{FCD$\downarrow$} & \multicolumn{1}{c}{Text2Mol$\uparrow$} & \multicolumn{1}{c}{Validity$\uparrow$} \\
\thickhline
Ground Truth & 100.0 & 100.0 & 0.00 & 100.0 & 100.0 & 100.0 & 0.00 & 11.26 & 100.0 \\
\hline
MolT5-Small & 56.56 & 0.00 & 56.34 & 64.22 & 58.10 & 37.44 & NaN & 0.49 & 80.52 \\
MolT5-Base & 68.38 & 0.00 & 44.79 & 76.03 & 65.23  & 47.46 & NaN & 7.06 & 100.0 \\
MolT5-Large & 56.42 & 0.00 & 55.40 & 75.70 & 65.01 & 39.51 & 17.52 & 7.69 & 99.44 \\
Meditron-7B & 69.40 & 0.01 & 46.49 & 77.16 & 69.34 & 50.07 & 2.46 & 7.80 & 99.63 \\
\end{tabular}
}
\caption{Molecule generation results on the validation split of \name{}. The FCD and Text2mol metrics are computed using only syntactically valid molecules. We found FCD suffers from numerical instability for the small and base models.
}
\label{tab:results_generation}
\end{table*}

\begin{table*}[ht!]
\resizebox{\textwidth}{!}{
\centering
\begin{tabular}{ c|cccccccccc }

\multicolumn{1}{c}{\textbf{Model}} & \multicolumn{1}{c}{BLEU$\uparrow$} & \multicolumn{1}{c}{Exact$\uparrow$} & \multicolumn{1}{c}{Levenshtein$\downarrow$} & \multicolumn{1}{c}{MACCS FTS$\uparrow$} & \multicolumn{1}{c}{RDK FTS$\uparrow$} & \multicolumn{1}{c}{Morgan FTS$\uparrow$} & \multicolumn{1}{c}{FCD$\downarrow$} & \multicolumn{1}{c}{Text2Mol$\uparrow$} & \multicolumn{1}{c}{Uniqueness$\uparrow$} & \multicolumn{1}{c}{Validity$\uparrow$} \\
\thickhline
Ground Truth & 100.0 & 100.0 & 0.00 & 100.0 & 100.0 & 100.0 & 0.0 & 23.05 & 100.0& 100.0 \\
\hline
MolT5-Small & 22.80 & 0.00 & 54.14 & 8.99 & 5.19 & 3.48 & NaN & 5.79 & 10.14 & 39.79 \\
MolT5-Base & 29.51 & 0.00 & 48.91 & 38.78 & 19.73 & 14.21 & NaN & 21.60 & 5.13 & 100.0 \\
MolT5-Large & 24.37 & 0.00 & 63.44 & 41.56 & 24.23 & 15.71 & NaN & 23.77 & 12.72 & 97.82 \\
Meditron-7B & 28.04 & 0.00 & 53.44 & 40.90 & 27.42 & 16.82 & 3.91 & 22.46 & 74.81 & 98.58 \\
\end{tabular}
}
\caption{Molecule generation results on the subset of held-out combinations from the validation split of \name{} (2107 data points). 
}
\label{tab:results_generation_heldout}
\end{table*}

\subsection{Splitting}
Duplicate molecules are merged using RDKit \cite{Landrum2021RDKit2021_03_2} and molecules which cannot be processed are removed. We split the data by first examining property combinations. 20\% of combinations are witheld into the evaluation set. From molecules in the remaining 80\%, we keep 80\% for training and put 20\% in evaluation. The evaluation set is split into two tasks: molecule captioning and molecule generation. For each task, only one modality will be released prior to the shared task results. 

The training set consists of 160,492 molecule-description pairs. For the evaluation set, both molecule generation and captioning contain 21,839 pairs. 
Further, special splits are released for the training set which allow for validation using the training data. They are constructed using the same procedure as the official evaluation dataset. 

\section{Evaluation Metrics}

Overall, we adopt the evaluation metrics proposed by \citet{edwards2022translation}. However, we include invalid molecules in the calculations of FTS metrics (setting the score to zero for invalid molecules). 
We also add a uniqueness metric to the generated molecules for held-out combinations of properties \cite{polykovskiy2020molecular}. 
Further, we also look at property-specific precision, recall, and F-1 scores. These scores are calculated by matching tokenized names in the generated captions. These scores are further aggregated across specific properties (e.g., inhibitors, X-icides, etc.) and the four broad categories. Aggregations are performed by averaging scores (i.e., macro-F1). We further compute these scores specifically for held-out combinations of properties. 

\section{Benchmarks}
MolT5 models \cite{edwards2022translation} were finetuned for 20 epochs on the ``split\_train'' data split and evaluated on the ``split\_valid'', both of which are available online. Huggingface's transformers \cite{wolf2019huggingface} was used for finetuning with a learning rate of 2e-5 and weight decay of 0.01. A batch size of 128 was used for small and base models, and a batch size of 48 for large models. Further, Meditron-7B \cite{chen2023meditron} was finetuned for 5 epochs with a context length of 930, 2e-6 learning rate, and batch size of 8/16 (molecule/caption generation). Models are released online. Results for captioning are reported in Tables \ref{tab:results_captioning}, \ref{tab:prop_results_captioning} and \ref{tab:subprop_results_captioning}. Tables \ref{tab:results_generation}, and \ref{tab:results_generation_heldout} shows results for molecule generation. 

\begin{figure*}[h!]
  \centering
  \includegraphics[width=1.0\textwidth]{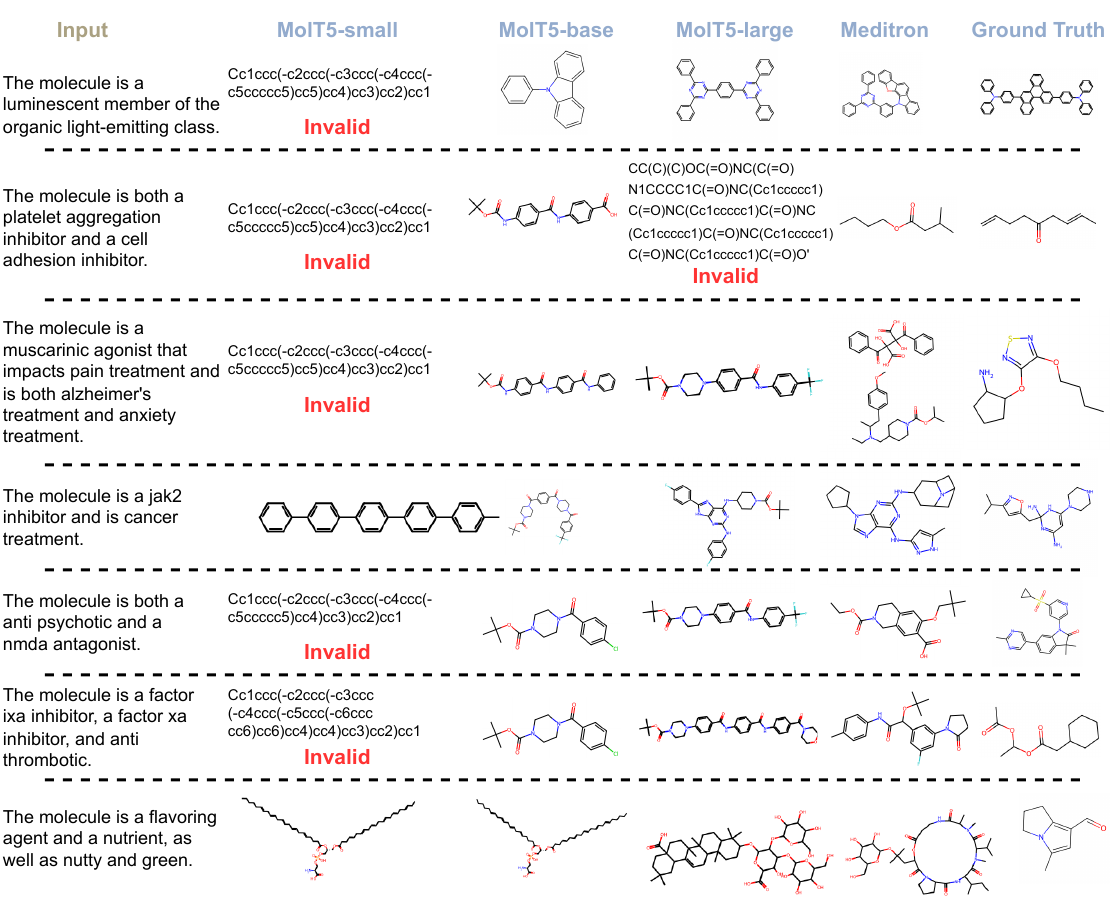}
\caption{ Examples of molecules generated by different models for never-before-seen property combinations. }
\label{fig:example1}
\end{figure*}

Overall, the dataset proves to be fairly challenging for these naively finetuned models. On captioning, Meditron-7B achieves a maximum overall F-1 score of 16.81 for property identification (Table \ref{tab:prop_results_captioning}). However, overall it has a much higher precision than recall, indicating the model only labels a molecule with a certain property when having higher confidence. Certain classes of molecules, such as X-icides, are never identified (Table \ref{tab:subprop_results_captioning}). Other classes, such as toxins or electricity, show emergent behavior as model size scales. Interestingly, the models appear to be fairly capable at linking molecules to certain diseases or cancers. We find that, likely due to poor performance on individual properties, only the largest model succeeds on predicting held-out combos, and with poor results. Additionally, we find that the Text2Mol metric, as trained on ChEBI-20, shows poor domain transfer to \name{}. For the final workshop evaluation, we plan to train an updated Text2Mol model on \name{}.

\begin{figure*}[h!]
  \centering
  \includegraphics[width=1.0\textwidth]{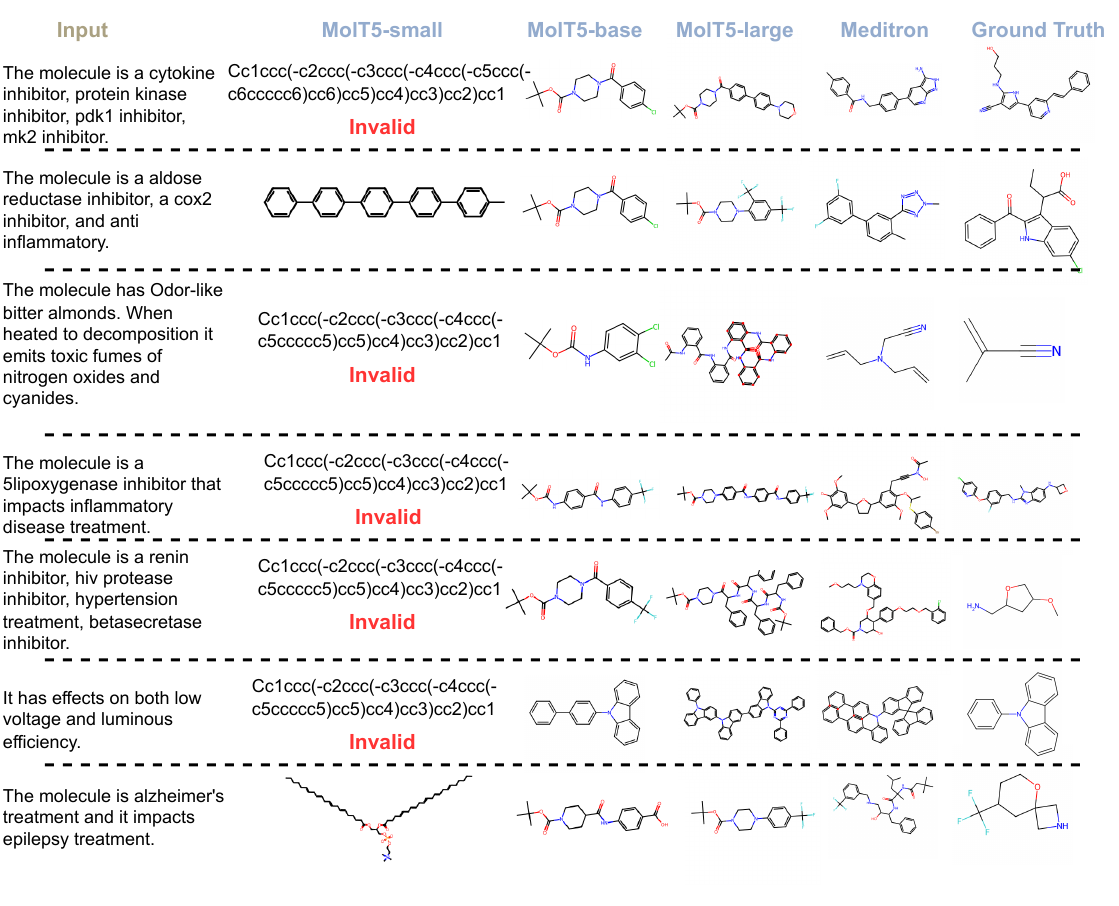}
\caption{ Examples of molecules generated by different models for never-before-seen property combinations. }
\label{fig:example2}
\end{figure*}
The models are able to capture a number of useful properties, such as electroluminescence, diabetes treatment, non-alcoholic fatty liver disease, and emulsifiers. In some cases, the model captures important characteristics about the molecule but uses differing language. This poses a challenge for our evaluation metrics. For example, a molecule identified in the ground truth as an anti tumor agent is identified as being a cancer treatment by the model. In particular, the models appear to struggle with rarer properties, which are common in our dataset formulation and in the chemical domain as a whole. They also struggle with identifying molecule-protein interactions (e.g., ``monoamine reuptake inhibitor''), although Meditron shows a large performance jump. 

For the molecule generation task, we also find the dataset to be challenging. We show results generated by different models on never-before-seen property combinations in Figures \ref{fig:example1} and \ref{fig:example2}. We believe the difficulty is for two reasons. First, common property combinations may have structurally very different molecules which exhibit those properties, making evaluation difficult. Second, the model may not grasp rare properties well. Overall, this results in the naively finetuned models producing similar outputs to many different prompts. Further, as expected, performance falls on unseen property combinations and larger models prove more effective (Table \ref{tab:results_generation_heldout}).

\section{Future Directions}
Overall, this dataset proves to be quite challenging. We find that some specific properties in particular are challenging for the model. This may be because the model understands these properties, but is unwilling to use them in its descriptions due to the training procedure. This limitation may be addressed with more sophisticated decoding algorithms or by better finetuning methods. Future work will also likely benefit from incorporating other modalities, such as proteins, to provide better understanding to the model for some property types. Notably, certain properties display what may be emergent behavior; scaling training data or model size may yield non-linear improvements.

In this dataset, we focus on composition, abstraction, and function. Future work may also wish to integrate other recent trends: instruction-following and dialogue \cite{fang2023mol,cao2023instructmol,zeng2023interactive,zhao2024chemdfm,zhang2024chemllm,yu2024llasmol}, tool use \cite{boiko2023emergent, bran2023chemcrow}, additional molecule representations (e.g., 3D \cite{tang2023mollm}), additional modalities \cite{xu2023multilingual}, or molecule editing \cite{su2022molecular}. Further, we note the need for improved evaluation metrics, especially in the case of molecule generation for function where there may be many possible outputs. Specific methods for improving compositionality may be another fruitful avenue for research \cite{yellinek20233vl}. 
It may also be interesting to use molecule-language instruction-following models within larger search frameworks, such as ChemReasoner \cite{sprueill2023monte, sprueill2024chemreasoner}.

\section{Conclusion}
In this manuscript, we describe the process for creating the \name{} dataset. \name{} is designed to focus on three key benefits of natural language in molecule design: compositionality, functionality, and abstraction. It will feature as the shared task at the \href{https://language-plus-molecules.github.io/}{First Language + Molecules Workshop} at ACL 2024. We encourage any submissions!

\section*{Acknowledgements}
We would like to thank the authors and creators of our data sources for freely allowing the use of their work. We would like to acknowledge NVIDIA Corporation’s contributions to this work through a grant of NVIDIA A100 Tensor Core GPUs. This work is supported by the Molecule Maker Lab Institute: an AI research institute program supported by NSF under award No. 2019897, by DOE Center for Advanced Bioenergy and Bioproducts Innovation U.S. Department of Energy, Office of Science, Office of Biological and Environmental Research under Award Number DESC0018420, and by AI Algriculture: the Agriculture and Food Research Initiative (AFRI) grant no. 2020-67021- 32799/project accession no.1024178 from the USDA National Institute of Food and Agriculture. The views and conclusions contained herein are those of the authors and should not be interpreted as necessarily representing the official policies, either expressed or implied of, the National Science Foundation, the U.S. Department of Energy, and the U.S. Government. The U.S. Government is authorized to reproduce and distribute reprints for governmental purposes notwithstanding any copyright annotation therein.

\bibliography{anthology,custom}

\begin{thebibliography}{43}
\expandafter\ifx\csname natexlab\endcsname\relax\def\natexlab#1{#1}\fi

\bibitem[{Bird et~al.(2009)Bird, Klein, and Loper}]{bird2009natural}
Steven Bird, Ewan Klein, and Edward Loper. 2009.
\newblock \emph{Natural language processing with Python: analyzing text with the natural language toolkit}.
\newblock " O'Reilly Media, Inc.".

\bibitem[{Boiko et~al.(2023)Boiko, MacKnight, and Gomes}]{boiko2023emergent}
Daniil~A Boiko, Robert MacKnight, and Gabe Gomes. 2023.
\newblock \href {https://arxiv.org/abs/2304.05332} {Emergent autonomous scientific research capabilities of large language models}.
\newblock \emph{ArXiv preprint}, abs/2304.05332.

\bibitem[{Bran et~al.(2023)Bran, Cox, White, and Schwaller}]{bran2023chemcrow}
Andres~M Bran, Sam Cox, Andrew~D White, and Philippe Schwaller. 2023.
\newblock Chemcrow: Augmenting large-language models with chemistry tools.
\newblock \emph{arXiv preprint arXiv:2304.05376}.

\bibitem[{Cao et~al.(2023)Cao, Liu, Lu, Yao, and Li}]{cao2023instructmol}
He~Cao, Zijing Liu, Xingyu Lu, Yuan Yao, and Yu~Li. 2023.
\newblock Instructmol: Multi-modal integration for building a versatile and reliable molecular assistant in drug discovery.
\newblock \emph{arXiv preprint arXiv:2311.16208}.

\bibitem[{Chen et~al.(2023{\natexlab{a}})Chen, Xi, Du, Wang, Jianyu, Zhao, and Qin}]{chen2023artificially}
Yuhan Chen, Nuwa Xi, Yanrui Du, Haochun Wang, Chen Jianyu, Sendong Zhao, and Bing Qin. 2023{\natexlab{a}}.
\newblock From artificially real to real: Leveraging pseudo data from large language models for low-resource molecule discovery.
\newblock \emph{arXiv preprint arXiv:2309.05203}.

\bibitem[{Chen et~al.(2023{\natexlab{b}})Chen, Cano, Romanou, Bonnet, Matoba, Salvi, Pagliardini, Fan, K{\"o}pf, Mohtashami et~al.}]{chen2023meditron}
Zeming Chen, Alejandro~Hern{\'a}ndez Cano, Angelika Romanou, Antoine Bonnet, Kyle Matoba, Francesco Salvi, Matteo Pagliardini, Simin Fan, Andreas K{\"o}pf, Amirkeivan Mohtashami, et~al. 2023{\natexlab{b}}.
\newblock Meditron-70b: Scaling medical pretraining for large language models.
\newblock \emph{arXiv preprint arXiv:2311.16079}.

\bibitem[{Christofidellis et~al.(2023)Christofidellis, Giannone, Born, Winther, Laino, and Manica}]{christofidellis2023unifying}
Dimitrios Christofidellis, Giorgio Giannone, Jannis Born, Ole Winther, Teodoro Laino, and Matteo Manica. 2023.
\newblock Unifying molecular and textual representations via multi-task language modelling.
\newblock \emph{arXiv preprint arXiv:2301.12586}.

\bibitem[{Edwards et~al.(2022)Edwards, Lai, Ros, Honke, Cho, and Ji}]{edwards2022translation}
Carl Edwards, Tuan Lai, Kevin Ros, Garrett Honke, Kyunghyun Cho, and Heng Ji. 2022.
\newblock \href {https://aclanthology.org/2022.emnlp-main.26} {Translation between molecules and natural language}.
\newblock In \emph{Proceedings of the 2022 Conference on Empirical Methods in Natural Language Processing}, pages 375--413, Abu Dhabi, United Arab Emirates. Association for Computational Linguistics.

\bibitem[{Edwards et~al.(2021)Edwards, Zhai, and Ji}]{edwards2021text2mol}
Carl Edwards, ChengXiang Zhai, and Heng Ji. 2021.
\newblock \href {https://doi.org/10.18653/v1/2021.emnlp-main.47} {{T}ext2{M}ol: Cross-modal molecule retrieval with natural language queries}.
\newblock In \emph{Proceedings of the 2021 Conference on Empirical Methods in Natural Language Processing}, pages 595--607, Online and Punta Cana, Dominican Republic. Association for Computational Linguistics.

\bibitem[{Fang et~al.(2023)Fang, Liang, Zhang, Liu, Huang, Chen, Fan, and Chen}]{fang2023mol}
Yin Fang, Xiaozhuan Liang, Ningyu Zhang, Kangwei Liu, Rui Huang, Zhuo Chen, Xiaohui Fan, and Huajun Chen. 2023.
\newblock Mol-instructions: A large-scale biomolecular instruction dataset for large language models.
\newblock \emph{arXiv preprint arXiv:2306.08018}.

\bibitem[{Ferguson and Gray(2018)}]{ferguson2018kinase}
Fleur~M Ferguson and Nathanael~S Gray. 2018.
\newblock Kinase inhibitors: the road ahead.
\newblock \emph{Nature reviews Drug discovery}, 17(5):353--377.

\bibitem[{Kim et~al.(2019)Kim, Chen, Cheng, Gindulyte, He, He, Li, Shoemaker, Thiessen, Yu et~al.}]{kim2019pubchem}
Sunghwan Kim, Jie Chen, Tiejun Cheng, Asta Gindulyte, Jia He, Siqian He, Qingliang Li, Benjamin~A. Shoemaker, Paul~A. Thiessen, Bo~Yu, et~al. 2019.
\newblock Pubchem 2019 update: improved access to chemical data.
\newblock \emph{Nucleic acids research}, 47(D1):D1102--D1109.

\bibitem[{Kim et~al.(2016)Kim, Thiessen, Bolton, Chen, Fu, Gindulyte, Han, He, He, Shoemaker et~al.}]{kim2016pubchem}
Sunghwan Kim, Paul~A. Thiessen, Evan~E. Bolton, Jie Chen, Gang Fu, Asta Gindulyte, Lianyi Han, Jane He, Siqian He, Benjamin~A. Shoemaker, et~al. 2016.
\newblock Pubchem substance and compound databases.
\newblock \emph{Nucleic acids research}, 44(D1):D1202--D1213.

\bibitem[{Kippelen and Br{\'e}das(2009)}]{kippelen2009organic}
Bernard Kippelen and Jean-Luc Br{\'e}das. 2009.
\newblock Organic photovoltaics.
\newblock \emph{Energy \& Environmental Science}, 2(3):251--261.

\bibitem[{Kosonocky et~al.(2023)Kosonocky, Wilke, Marcotte, and Ellington}]{kosonocky2023mining}
Clayton~W Kosonocky, Claus~O Wilke, Edward~M Marcotte, and Andrew~D Ellington. 2023.
\newblock Mining patents with large language models demonstrates congruence of functional labels and chemical structures.
\newblock \emph{arXiv preprint arXiv:2309.08765}.

\bibitem[{Landrum(2021)}]{Landrum2021RDKit2021_03_2}
Greg Landrum. 2021.
\newblock \href {https://github.com/rdkit/rdkit/releases/tag/Release_2021_03_2} {Rdkit: Open-source cheminformatics software}.

\bibitem[{Liu et~al.(2023{\natexlab{a}})Liu, Ren, and Ren}]{liu2023git}
Pengfei Liu, Yiming Ren, and Zhixiang Ren. 2023{\natexlab{a}}.
\newblock Git-mol: A multi-modal large language model for molecular science with graph, image, and text.
\newblock \emph{arXiv preprint arXiv:2308.06911}.

\bibitem[{Liu et~al.(2022)Liu, Nie, Wang, Lu, Qiao, Liu, Tang, Xiao, and Anandkumar}]{liu2022multi}
Shengchao Liu, Weili Nie, Chengpeng Wang, Jiarui Lu, Zhuoran Qiao, Ling Liu, Jian Tang, Chaowei Xiao, and Anima Anandkumar. 2022.
\newblock Multi-modal molecule structure-text model for text-based retrieval and editing.
\newblock \emph{arXiv preprint arXiv:2212.10789}.

\bibitem[{Liu et~al.(2023{\natexlab{b}})Liu, Zhang, Xia, Wu, Xie, Qin, Zhang, and Liu}]{liu2023molxpt}
Zequn Liu, Wei Zhang, Yingce Xia, Lijun Wu, Shufang Xie, Tao Qin, Ming Zhang, and Tie-Yan Liu. 2023{\natexlab{b}}.
\newblock Molxpt: Wrapping molecules with text for generative pre-training.
\newblock \emph{arXiv preprint arXiv:2305.10688}.

\bibitem[{Liu et~al.(2023{\natexlab{c}})Liu, Li, Luo, Fei, Cao, Kawaguchi, Wang, and Chua}]{liu2023molca}
Zhiyuan Liu, Sihang Li, Yanchen Luo, Hao Fei, Yixin Cao, Kenji Kawaguchi, Xiang Wang, and Tat-Seng Chua. 2023{\natexlab{c}}.
\newblock Molca: Molecular graph-language modeling with cross-modal projector and uni-modal adapter.
\newblock \emph{arXiv preprint arXiv:2310.12798}.

\bibitem[{Luo et~al.(2023)Luo, Yang, Hong, Liu, and Nie}]{luo2023molfm}
Yizhen Luo, Kai Yang, Massimo Hong, Xingyi Liu, and Zaiqing Nie. 2023.
\newblock Molfm: A multimodal molecular foundation model.
\newblock \emph{arXiv preprint arXiv:2307.09484}.

\bibitem[{OpenAI(2023)}]{openai2023gpt4}
OpenAI. 2023.
\newblock \href {https://arxiv.org/abs/2303.08774} {Gpt-4 technical report}.
\newblock \emph{ArXiv preprint}, abs/2303.08774.

\bibitem[{Pei et~al.(2023)Pei, Zhang, Zhu, Wu, Gao, Wu, Xia, and Yan}]{pei2023biot5}
Qizhi Pei, Wei Zhang, Jinhua Zhu, Kehan Wu, Kaiyuan Gao, Lijun Wu, Yingce Xia, and Rui Yan. 2023.
\newblock Biot5: Enriching cross-modal integration in biology with chemical knowledge and natural language associations.
\newblock \emph{arXiv preprint arXiv:2310.07276}.

\bibitem[{Polykovskiy et~al.(2020)Polykovskiy, Zhebrak, Sanchez-Lengeling, Golovanov, Tatanov, Belyaev, Kurbanov, Artamonov, Aladinskiy, Veselov et~al.}]{polykovskiy2020molecular}
Daniil Polykovskiy, Alexander Zhebrak, Benjamin Sanchez-Lengeling, Sergey Golovanov, Oktai Tatanov, Stanislav Belyaev, Rauf Kurbanov, Aleksey Artamonov, Vladimir Aladinskiy, Mark Veselov, et~al. 2020.
\newblock Molecular sets (moses): a benchmarking platform for molecular generation models.
\newblock \emph{Frontiers in pharmacology}, 11:1931.

\bibitem[{Seidl et~al.(2023)Seidl, Vall, Hochreiter, and Klambauer}]{seidl2023enhancing}
Philipp Seidl, Andreu Vall, Sepp Hochreiter, and G{\"u}nter Klambauer. 2023.
\newblock \href {https://arxiv.org/abs/2303.03363} {Enhancing activity prediction models in drug discovery with the ability to understand human language}.
\newblock \emph{ArXiv preprint}, abs/2303.03363.

\bibitem[{Sprueill et~al.(2024)Sprueill, Edwards, Agarwal, Olarte, Sanyal, Johnston, Liu, Ji, and Choudhury}]{sprueill2024chemreasoner}
Henry~W Sprueill, Carl Edwards, Khushbu Agarwal, Mariefel~V Olarte, Udishnu Sanyal, Conrad Johnston, Hongbin Liu, Heng Ji, and Sutanay Choudhury. 2024.
\newblock Chemreasoner: Heuristic search over a large language model's knowledge space using quantum-chemical feedback.
\newblock \emph{arXiv preprint arXiv:2402.10980}.

\bibitem[{Sprueill et~al.(2023)Sprueill, Edwards, Olarte, Sanyal, Ji, and Choudhury}]{sprueill2023monte}
Henry~W Sprueill, Carl Edwards, Mariefel~V Olarte, Udishnu Sanyal, Heng Ji, and Sutanay Choudhury. 2023.
\newblock Monte carlo thought search: Large language model querying for complex scientific reasoning in catalyst design.
\newblock \emph{arXiv preprint arXiv:2310.14420}.

\bibitem[{Su et~al.(2022)Su, Du, Yang, Zhou, Li, Rao, Sun, Lu, and Wen}]{su2022molecular}
Bing Su, Dazhao Du, Zhao Yang, Yujie Zhou, Jiangmeng Li, Anyi Rao, Hao Sun, Zhiwu Lu, and Ji-Rong Wen. 2022.
\newblock \href {https://arxiv.org/abs/2209.05481} {A molecular multimodal foundation model associating molecule graphs with natural language}.
\newblock \emph{ArXiv preprint}, abs/2209.05481.

\bibitem[{Tang et~al.(2023)Tang, Tran, Tan, and Gerstein}]{tang2023mollm}
Xiangru Tang, Andrew Tran, Jeffrey Tan, and Mark~B Gerstein. 2023.
\newblock Mollm: A unified language model to integrate biomedical text with 2d and 3d molecular representations.
\newblock \emph{bioRxiv}, pages 2023--11.

\bibitem[{Wishart et~al.(2023)Wishart, Girod, Peters, Oler, Jovel, Budinski, Milford, Lui, Sayeeda, Mah et~al.}]{wishart2023chemfont}
David~S Wishart, Sagan Girod, Harrison Peters, Eponine Oler, Juan Jovel, Zachary Budinski, Ralph Milford, Vicki~W Lui, Zinat Sayeeda, Robert Mah, et~al. 2023.
\newblock Chemfont: the chemical functional ontology resource.
\newblock \emph{Nucleic Acids Research}, 51(D1):D1220--D1229.

\bibitem[{Wolf et~al.(2019)Wolf, Debut, Sanh, Chaumond, Delangue, Moi, Cistac, Rault, Louf, Funtowicz et~al.}]{wolf2019huggingface}
Thomas Wolf, Lysandre Debut, Victor Sanh, Julien Chaumond, Clement Delangue, Anthony Moi, Pierric Cistac, Tim Rault, R{\'e}mi Louf, Morgan Funtowicz, et~al. 2019.
\newblock Huggingface's transformers: State-of-the-art natural language processing.
\newblock \emph{arXiv preprint arXiv:1910.03771}.

\bibitem[{Xu et~al.(2023)Xu, Woicik, Poon, Altman, and Wang}]{xu2023multilingual}
Hanwen Xu, Addie Woicik, Hoifung Poon, Russ~B Altman, and Sheng Wang. 2023.
\newblock Multilingual translation for zero-shot biomedical classification using biotranslator.
\newblock \emph{Nature Communications}, 14(1):738.

\bibitem[{Yellinek et~al.(2023)Yellinek, Karlinsky, and Giryes}]{yellinek20233vl}
Nir Yellinek, Leonid Karlinsky, and Raja Giryes. 2023.
\newblock 3vl: using trees to teach vision \& language models compositional concepts.
\newblock \emph{arXiv preprint arXiv:2312.17345}.

\bibitem[{Yu et~al.(2024)Yu, Baker, Chen, Ning, and Sun}]{yu2024llasmol}
Botao Yu, Frazier~N Baker, Ziqi Chen, Xia Ning, and Huan Sun. 2024.
\newblock Llasmol: Advancing large language models for chemistry with a large-scale, comprehensive, high-quality instruction tuning dataset.
\newblock \emph{arXiv preprint arXiv:2402.09391}.

\bibitem[{Zeng et~al.(2023)Zeng, Yin, Wang, Liu, Yang, Yao, Sun, Sun, Xie, and Liu}]{zeng2023interactive}
Zheni Zeng, Bangchen Yin, Shipeng Wang, Jiarui Liu, Cheng Yang, Haishen Yao, Xingzhi Sun, Maosong Sun, Guotong Xie, and Zhiyuan Liu. 2023.
\newblock Interactive molecular discovery with natural language.
\newblock \emph{arXiv preprint arXiv:2306.11976}.

\bibitem[{Zeng et~al.(2022)Zeng, Li, Gasevic, and Chen}]{zeng-etal-2022-deep}
Zijie Zeng, Xinyu Li, Dragan Gasevic, and Guanliang Chen. 2022.
\newblock \href {https://doi.org/10.18653/v1/2022.naacl-main.14} {Do deep neural nets display human-like attention in short answer scoring?}
\newblock In \emph{Proceedings of the 2022 Conference of the North American Chapter of the Association for Computational Linguistics: Human Language Technologies}, pages 191--205, Seattle, United States. Association for Computational Linguistics.

\bibitem[{Zhang et~al.(2024)Zhang, Liu, Tan, Chen, Yan, Yan, Li, Huang, Yue, Zhou et~al.}]{zhang2024chemllm}
Di~Zhang, Wei Liu, Qian Tan, Jingdan Chen, Hang Yan, Yuliang Yan, Jiatong Li, Weiran Huang, Xiangyu Yue, Dongzhan Zhou, et~al. 2024.
\newblock Chemllm: A chemical large language model.
\newblock \emph{arXiv preprint arXiv:2402.06852}.

\bibitem[{Zhang et~al.(2023)Zhang, Wang, Helwig, Luo, Fu, Xie, Liu, Lin, Xu, Yan et~al.}]{zhang2023artificial}
Xuan Zhang, Limei Wang, Jacob Helwig, Youzhi Luo, Cong Fu, Yaochen Xie, Meng Liu, Yuchao Lin, Zhao Xu, Keqiang Yan, et~al. 2023.
\newblock Artificial intelligence for science in quantum, atomistic, and continuum systems.
\newblock \emph{arXiv preprint arXiv:2307.08423}.

\bibitem[{Zhao et~al.(2023{\natexlab{a}})Zhao, Liu, Ma, Xu, Fu, Deng, Kong, and Liu}]{zhao2023gimlet}
Haiteng Zhao, Shengchao Liu, Chang Ma, Hannan Xu, Jie Fu, Zhi-Hong Deng, Lingpeng Kong, and Qi~Liu. 2023{\natexlab{a}}.
\newblock Gimlet: A unified graph-text model for instruction-based molecule zero-shot learning.
\newblock \emph{bioRxiv}, pages 2023--05.

\bibitem[{Zhao et~al.(2023{\natexlab{b}})Zhao, Edwards, and Ji}]{zhao2023scientific}
Lawrence Zhao, Carl Edwards, and Heng Ji. 2023{\natexlab{b}}.
\newblock What a scientific language model knows and doesn't know about chemistry.
\newblock In \emph{NeurIPS 2023 AI for Science Workshop}.

\bibitem[{Zhao et~al.(2023{\natexlab{c}})Zhao, Zhou, Cao, Zhang, and Chen}]{zhao2023adversarial}
Wenyu Zhao, Dong Zhou, Buqing Cao, Kai Zhang, and Jinjun Chen. 2023{\natexlab{c}}.
\newblock Adversarial modality alignment network for cross-modal molecule retrieval.
\newblock \emph{IEEE Transactions on Artificial Intelligence}.

\bibitem[{Zhao et~al.(2024)Zhao, Ma, Chen, Sun, Li, Xu, Zhu, Zhu, Fan, Shen et~al.}]{zhao2024chemdfm}
Zihan Zhao, Da~Ma, Lu~Chen, Liangtai Sun, Zihao Li, Hongshen Xu, Zichen Zhu, Su~Zhu, Shuai Fan, Guodong Shen, et~al. 2024.
\newblock Chemdfm: Dialogue foundation model for chemistry.
\newblock \emph{arXiv preprint arXiv:2401.14818}.

\bibitem[{Zhong et~al.(2023)Zhong, Ouyang, Jiao, Kargupta, Luo, Shen, Zhou, Zhong, Liu, Li, Xiao, Jiang, Hu, Wang, Ji, Burke, Zhao, and Han}]{reactionminer2023}
Ming Zhong, Siru Ouyang, Yizhu Jiao, Priyanka Kargupta, Leo Luo, Yanzhen Shen, Bobby Zhou, Xianrui Zhong, Xuan Liu, Hongxiang Li, Jinfeng Xiao, Minhao Jiang, Vivian Hu, Xuan Wang, Heng Ji, Martin Burke, Huimin Zhao, and Jiawei Han. 2023.
\newblock Reaction miner: An integrated system for chemical reaction extraction from textual data.
\newblock In \emph{Proc. The 2023 Conference on Empirical Methods in Natural Language Processing (EMNLP2023) Demo Track}.

\end{thebibliography}
\bibliographystyle{acl_natbib}

\appendix

\section{Prompts and examples for GPT4}
\begin{itemize}
    \item Prompts: You are an expert in the chemical domain whose task is to create templates to describe the properties of molecules. You will be challenged with a list of different cases. Each case willl have a list of **templates**, and a **question**. Each template will describe certain properties. Your goal is to generate a new template in a sentence based on  all the previous templates.
    \item Case 1: Templates: - The molecule is a <0>. - It belongs to the <1> class of molecules. Answer: The molecule, characterized as a <0>, falls under the <1> category of chemical compounds. 
    \item Case 2: Templates: - It has an effect on <2>. - It impacts <3>. Answer: It impacts <2> and has an effect on <3>. 
    \item Case 3: Templates: - The molecule is <4>. - The molecule has a <5>. Answer: The molecule is <4>. and has a <5>. 
    \item Case 4: Templates: - The molecule is a <0\_1>. - The molecule is a <0\_2>. Answer: The molecule is a <0\_1> and exhibits <0\_2> properties. 
    \item Case 5: Templates: - It belongs to the <1\_1> class of molecules. - It belongs to the <1\_2> class of molecules. Answer: The molecule is in the <1\_1> class of compounds, characterizing it as a member of the <1\_2> family. 
\end{itemize}

\section{Additional Dataset Statistics}
\begin{figure*}[h!]
  \centering
  \includegraphics[width=1.0\textwidth]{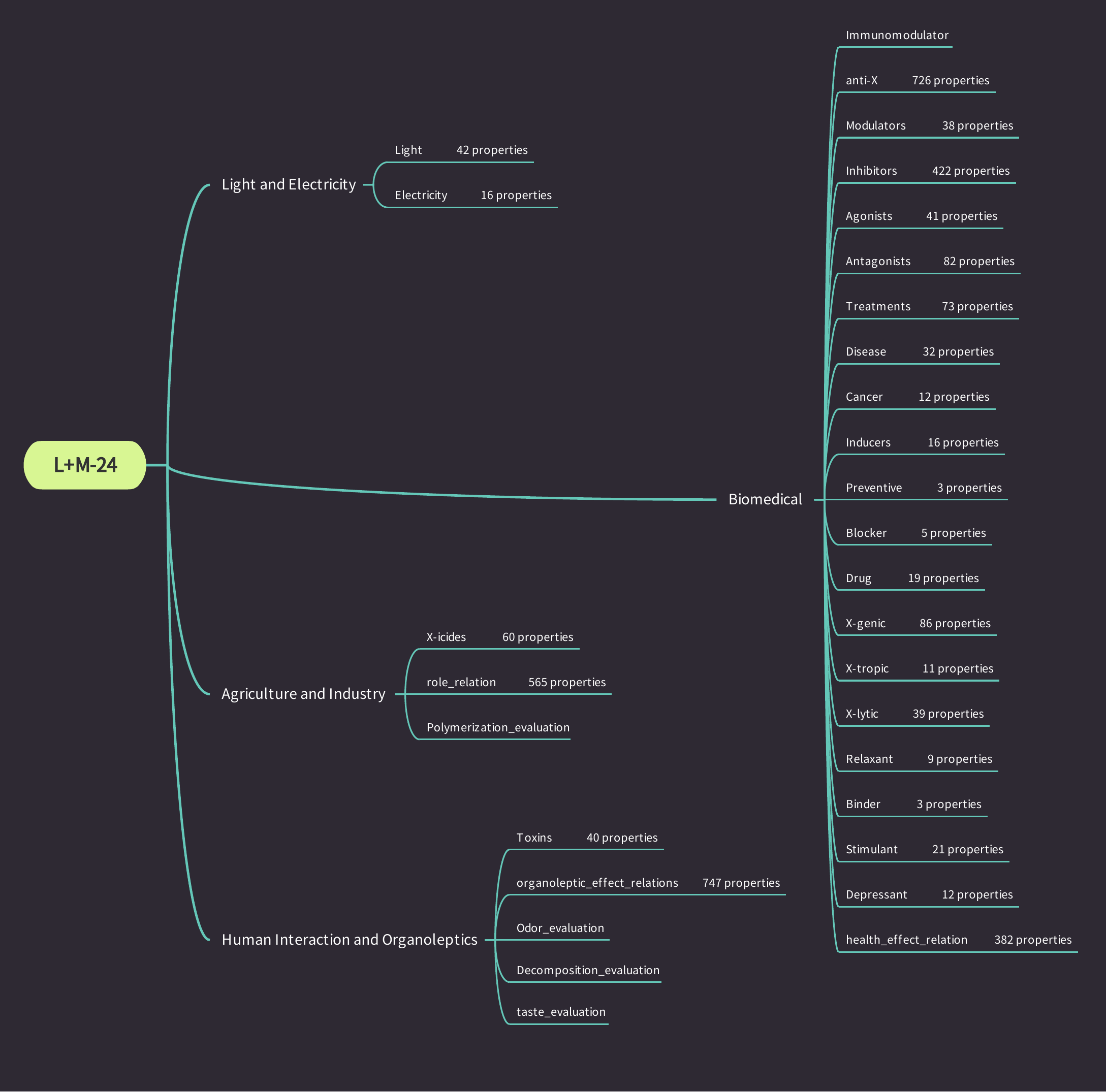}
\caption{ Breakdown of different property classes in \name{}. }
\label{fig:property_breakdown}
\end{figure*}

Here, we give a brief description of properties in the dataset. Table \ref{tab:data_stats} shows the number of property-molecule pairs for different property classes. Figure \ref{fig:property_breakdown} breaks the dataset down into different property classes. More details can be found in the dataset repository. 

\begin{table}[ht!]
\resizebox{\columnwidth}{!}{

\centering
\begin{tabular}{ cc }

Group & Property-Molecule Pair Count \\
\hline
\textbf{Total} & 1512865 \\
\hline
\textbf{Biomedical} & 776712 \\
anti-X & 24884 \\
Modulators & 2787 \\
Inhibitors & 23257 \\
Agonists & 1161 \\
Antagonists & 3172 \\
 Treatments & 53070 \\
 Disease & 316380 \\
 Cancer & 41456 \\
 Inducers & 31 \\
 Preventive & 0 \\
 Blocker & 47 \\
 Drug & 260 \\
 X-genic & 172 \\
 X-tropic & 17 \\
 X-lytic & 84 \\
 Relaxant & 40 \\
 Binder & 4 \\
 Stimulant & 60 \\
 Depressant & 52 \\
 health\_effect\_relations & 309532 \\
\hline
\textbf{Light and Electricity} & 14077 \\
 Light & 11069 \\
 Electricity & 3008 \\
\hline
\textbf{Human Interaction} & 27457 \\
 Toxins & 1070 \\
 organoleptic\_effect\_relations & 20501 \\
\hline
\textbf{Agric. and Industry} & 694619 \\
 X-icides & 809 \\
 role\_relation & 693648 \\

\end{tabular}
}
\caption{Number of property-molecule pairs for different property groups.
}
\label{tab:data_stats}
\end{table}

\end{document}